\documentclass[letterpaper, 10 pt, conference]{ieeeconf}  

\IEEEoverridecommandlockouts                              
\usepackage{listings}
\usepackage{framed}
\usepackage{cite} 
\usepackage{forest}
\usepackage{algorithm, algpseudocode}
\usepackage{graphicx}
\usepackage{caption}
\usepackage{subcaption}

\title{\LARGE \bf
Scalable, Trie-based Approximate Entity Extraction for Real-Time Financial Transaction Screening
}

\author{ 
\parbox{3 in}{\centering Emrah Budur 
         \\
         Garanti Technology\\
		 34212, Istanbul, Turkey\\
         {\tt\small emrahbu [at] garanti.com.tr}}
         \hspace*{ 0.5 in}
}

\forestset{
  sn edges/.style={for tree={parent anchor=south, child anchor=north,align=center,base=top}, for tree={where n children=0{tier=word, delay=with translation}{}}},
  trace/.style={
    replace by={[\trace, delay={}, tier=word]}
  },
  with translation/.style={
    l sep=0,
    inner xsep=0,
    append translation/.expanded/.wrap pgfmath arg={\gettranslation{##1}}{content},
    content/.expanded/.wrap pgfmath arg={\gettext{##1}}{content},
  },
  append translation/.style={
    append={
      [
        #1,
        no edge,
        l=0,
        inner xsep=0,
        inner ysep=0,
        outer ysep=0,
        before computing xy={
          l-=2pt
        }
      ]
    },
  },
}

\def\gettext#1{\gettextA#1;;\endget}
\def\gettextA#1;#2;{\removesep#1;}
\def\gettranslation#1{\gettranslationA#1;;\endget}
\def\gettranslationA#1;#2;{\removesep#2;}
\def\removesep#1;#2\endget{#1}

\newcommand{\trace}{\raisebox{0.2ex}{\_}\rule{0cm}{0.7em}}

\begin{document}

\maketitle
\thispagestyle{empty}
\pagestyle{empty}

\begin{abstract}
Financial institutions have to screen their transactions to ensure that they are not affiliated with terrorism entities.  Developing appropriate solutions to detect such affiliations precisely while avoiding any kind of interruption to large amount of legitimate transactions is essential.  In this paper, we present building blocks of a scalable solution that may help financial institutions to build their own software to extract terrorism entities out of both structured and unstructured financial messages in real time and with approximate similarity matching approach.
\end{abstract}

\section{Introduction}

In September 11, 2011, one of the deadliest terrorism attack in US happened.  After this event, the US government strengthened the regulation against terrorism financing. The most important responsibility was given to the financial institutions, worldwide.  Financial institutions have to screen  their transactions against terrorism entities and block any transaction affiliated with these entities.  These entities are published\footnote{They  can be collected  either freely from the web or on a licence basis from some private institutions including Thomson Reuters \cite{2016World-Check}} by the governmental institutions including the Office of Foreign Assets Control (OFAC)  \cite{2016SpeciallySDN}.

Let's give a concrete example for what kind of responsibility that financial institutions have.  Figure \ref{fig:sample_message} is a typical swift message that is used for international money transfers.

\begin{figure}
\begin{framed}
\begin{lstlisting}
{1:F01MIDLGB22AXXX0548034693}
{2:I103BKTRUS33XBRDN3}
{3:{108:MT103}}
{4:
:20:8861198-0706
:23B:CRED
:32A:000612USD5443,99
:33B:USD5443,99
:50K:MIYESE INTERNATIONAL LIMITED
:52A:BCITITMM500
:53A:BCITUS33
:54A:IRVTUS3N
:57A:BNPAFRPPGRE
:59:/20041010050500001M02606
AHMET EMRE 
:70:/RFB/OYA/INVOICE SENT (*\bfseries TAMERLAAN \\
TZARNAEV*), FATIME ST. PLAZA DE HALIT
28934 MOSTOLES (MADRID)
:71A:SHA
-}
\end{lstlisting}
\end{framed}
  \captionof{figure}{Typical swift message}
  \label{fig:sample_message}
\end{figure}

The problem with this message is that the bold name, \textbf{TAMERLAAN TZARNAEV}, is in a watch list of the suspected international terrorists, namely TIDE \cite{2016TerroristEnvironment}, which is maintained by the well-known intelligence institutions such as CIA, FBI and NSA.  This kind of affiliation with terrorism entities in the financial messages should be detected and the message should be immediately blocked.  Financial institutions will be on the hook for complicity aiding and abetting a terrorism action by failing to detect this kind of affiliation with terrorism entities.  However, the identification of the same entity, \textit{TAMERLAAN TZARNAEV}, in a similar context failed by the US Customs Authorities which led up to Boston Bombing afterwards \cite{Schmitt2016TamerlanLists}.  Hence, achieving precise detection of such critical entities is a challenging issue.

As exemplified above, the detection of entities out of unstructured text poses some challenges that are listed below.

\begin{itemize}
\item \textbf{Name variations:} Terrorism entities deliberately change their names a lot.  Hence, the list of names of terrorism entities is getting bigger and bigger.  The algorithm is supposed to cover all variations of the names of entities.  So, the challenge of detecting names is getting tougher.

\item \textbf{Fault tolerant match:} Even if the exact name of the entity is known to the application, it is possible that the name is misspelled in the transaction.  So, the algorithm is supposed to tolerate the misspellings which presents additional challenge.

\item \textbf{Mining unstructured text:} The names are not given in a clean structured field.  Instead, it is given as an unstructured field, i.e. explanation text, and the field may or may not include the name in it, also it may or may not include irrelevant data such as an address, which makes it harder to extract the relevant information precisely.

\item \textbf{Noisy words:} There are some terms which are frequently occurring both in illegal entities and legitimate entities, i.e. LIMITED.  The algorithm needs to take this situation into account  and avoid blocking legitimate entities due to a common term.

\item \textbf{Minimizing false positives:} Blocking a legitimate transaction is a false alert which threatens profitability of the financial institution.  For example, the algorithm is supposed to match the text “Hosein” with the name “Hussein” while it should avoid blocking the text “Saydam” due to its close textual proximity to the name “Saddam”. Hence, the algorithm is supposed to avoid false alerts as much as possible.

\item \textbf{Avoiding false negatives:} Any entities which are involved in the query either exactly or approximately should be extracted uncompromisingly.

\item \textbf{Ensuring low latency:} The algorithm needs to complete checking the transaction in subseconds in order not to interrupt the business workflows. 
\end{itemize}

The algorithm that will be presented in this paper is designed and implemented to meet all of these constraints.

\section{Related Work}
Approximate entity extraction has drawn much attention by researchers \cite{Wang2009EfficientConstraints,Chakrabarti2008AnChecking, Deng2012AnConstraints, Deng2014AExtraction}.
It was reported to be useful to extract approximate product names from product review articles \cite{Chakrabarti2008AnChecking, Deng2014AExtraction},  author names and paper titles from publication records \cite{Chakrabarti2008AnChecking,Deng2012AnConstraints} gene and protein lexicon from publication records \cite{Wang2009EfficientConstraints}.  One of the few relevant studies in financial industry was suggested by Xu et al. to extract corporate entities out of free format financial contracts  \cite{Xu2016ExploitingDocuments}.  In this paper, we will present an entity extraction framework and analyze its efficiency by extracting terrorism entities out of unstructured financial messages.

Various methods were presented by researchers to accommodate several  problems of the task. One of the problems is the fact that the number of all possible typing errors, which is potentially embedded in the query text, grows exponentially with respect to the number of allowed typing errors.  As a common approach, all of the possible typing errors in query is fabricated and probed against the index. However, most of the fabricated query terms doesn't occur at all in the target dictionary hence probing inexistent terms makes the algorithms inefficient and unscalable.  A number researchers proposed to limit the number of allowed errors to 1-edit, to minimize the number of probing  \cite{Cisak2015AMismatches, Brodal1996ApproximateQueries}.  However, this approach fails to identify the k-edit error matches where $k>1$, for example the match of the query term \textit{Hosein} with the record term \textit{Hussain} where $k=3$.  Although over \%80 of the errors are estimated to be 1-edit errors\cite{Cisak2015AMismatches}, maximizing extraction recall is essential when detecting terrorism entities in financial industry.  Hence, we proposed a scalable algorithm that covers k-edit errors for reasonably large $k$ while probing no inexistent term at all.  

Another common problem is to find a suitable indexing scheme for the target dictionary dataset that will help addressing fault tolerant matching, scalability and low latency constraints.  A common approach for this problem is to exploit q-gram indexing to address fault tolerant search constraint  \cite{Li2011FaerieExtraction,Cisak2015AMismatches,Deng2014AExtraction,Deng2012AnConstraints,Zobel1995FindingLexicons}.  However, there is a problem with this approach which was emphasized also by Wang et al. \cite{Wang2009EfficientConstraints}.  The length of the q-gram is bounded by the smallest term in the target dictionary dataset.  This leads to short length q-grams which result in long posting lists.  This situation makes the solution unscalable and slow since merging long posting lists is time and CPU intensive operation.  The problem was justified also by Zobel et al. and Xiao et al. \cite{Zobel1995FindingLexicons,Xiao2008Ed-JoinConstraints}.  Our algorithm minimizes the problem by indexing terms as a whole instead of q-grams while ensuring that scalability, low latency and fault tolerant search constraints are addressed.

Many researchers proposed solutions based on common dissimilarity measures such as Edit Distance, Hamming Distance, Jaccard Distance \cite{Wang2009EfficientConstraints,Deng2012AnConstraints,Xiao2008Ed-JoinConstraints,Deng2014AExtraction}.  On the other hand a few researchers proposed probabilistic hash-based solutions such as locality sensitive hashing (LSH) \cite{Gionis1999SimilarityHashing}.  However, LSH approach violates the \textit{avoiding false negatives} constraint due to the fact that LSH may miss some true positive results  \cite{Chakrabarti2008AnChecking}.  Therefore, we decided to proceed with a non-LSH approach, namely edit distance similarity approach, to avoid probabilistic false negative matches.  We left other kind of similarity metrics such as Hamming Distance, Jaccard Distance as a future work.

Edit distance can be implemented in various forms.  Following Ukonnen, many researchers adopted q-gram based method \cite{Ukkonen1992ApproximateMatches}. In q-gram based method, matching q-grams of two strings are considered to reflect the similarity of these strings.  However, this approach suffers from long posting list problem as described previously.  An alternative form of edit distance implementation is to employ trie data structure as explained by Arslan and Egecioglu\cite{ARSLAN2004DictionaryDistance}. Trie-based approach eliminates the long posting list problem of q-gram based implementation.
Hence, we used trie data structure for indexing.

\section{PROBLEM DEFINITION}

\subsection{Notation}
Let $\Sigma$ denotes the alphabet.  Let D denote the set of documents in target dictionary. Let $q$ denotes a tokenizable free text query. Let $d \in D$ denote a tokenizable document.  Let $t_q$ and $t_d$ denotes tokens in $q$ and $d$ respectively. Let $\|t_q\|$ and $\|t_d\|$ denote the length of the tokens $t_q$ and $t_d$ respectively. 

Let the function $DocFreq(t_d)$ return the document frequency of the record token $t_d$ in the target dictionary set $D$. Let the function $TF(t_d)$ return the term frequency while the function $IDF(t_d)$ return the inverse document frequency of the record token $t_d$.  Let the function $I(t_d)=TF(t_d) \times IDF(t_d)$ gives the information of the record token $t_d$.

Let the function $Edit(t_q, t_d)$ gives the edit distance and $Edit^w(t_q, t_d)$ gives the weighted edit distance between the tokens $t_q$ and $t_d$.  

Let $\tau_l$ refers to predefined edit distance threashold which is applied to a token of length $l$.  

Let the function 

\begin{equation}
Sim(t_q, t_d) = 1- \frac{Edit(t_q, t_d)}{max(\|t_q\|, \|t_d\|)}
\end{equation}

 gives the edit similarity while
 
 \begin{equation}
Sim^w(t_q, t_d) = 1- \frac{Edit^w(t_q, t_d)}{max(\|t_q\|, \|t_d\|)}
\end{equation}

 gives weighted edit similarity between the tokens $t_q$ and $t_d$.  

Let \textit{a match}  
 \begin{equation}
m_d = (t_q, t_d, e)
 \end{equation}
refers to a matched token pairs $(t_q, t_d)$  in the query $q$ where $Edit(t_q, t_d) = e < \epsilon$.

Given a query $q$ and a document $d$, a set of all possible matches $m_d = (t_q, t_d, e)$ is denoted as \textit{a match set} $M_d$ where $m_d \in M_d$. The function $Support(M_d)$ return the number of documents that contain all record tokens $t_d$ in $M_d$.

Given a document $d$ and a corresponding match set $M_d$  let 
 \begin{equation}
0 \leq R(M_d, d) \leq 100
 \end{equation}
be a ranking function for document $d \in D$ against the match set $M_d$ where the output reflects how approximately the document $d$ is involved in the query $q$. Note that $R(M_d,d)=100$ refers to exact match while $R(M_d,d)=0$ refers to no match. 

Let $\sigma$ refers to predefined percentage score threashold that is used to filter out the acceptable candidate sets.

\subsection{Problem Formulation}

Given a query $q$, we aim to extract top $k$ documents $d$ from $q$ such that $R(M_d, d) \geq \sigma$.  

\section{SYSTEM ARCHITECTURE}

\begin{figure}[hb]
  \centering
  \includegraphics[width=3.5in]{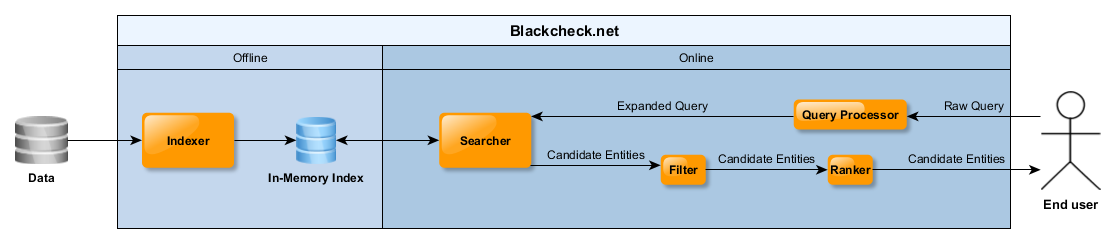}
  \caption[System overview diagram]
   {System architecture diagram}
  \label{figure:system_overview}
\end{figure}

The main building blocks of the application we present in this paper  are illustrated in Figure \ref{figure:system_overview} as a system architecture diagram. The details of the modules illustrated in the diagram are explained in subsequent sections.
 
The lifecycle of the application can be reviewed in two phases, namely offline and online.  In the offline phase, the target dictionary of sanctioned entities is indexed and loaded into main memory for later access.  In the online phase, the queries are processed and searched from the in-memory index while the candidate matches are returned as a result, in near real-time.

\subsection{Indexing} 

In this step, the list of target dictionary entities are indexed and loaded into main memory for later access. Although use of trie data structure was discouraged by Zobel et al.\cite{Zobel1995FindingLexicons} due to space complexity, our application showed that the advantages gained in terms of time complexity outweighs the disadvantages coming from space complexity.  Hence, we implemented trie data structure as suggested by Arslan and Egecioglu \cite{ARSLAN2004DictionaryDistance}.  

\subsubsection{Trie Index}

The trie data structure was originally proposed by Briandais  \cite{DeLaBriandais1959FileKeys}.  Later on, it was named by Fredkin \cite{Fredkin1960TrieMemory} reflecting the word re\textbf{\textit{trie}}val.  

\begin{figure*}[h]
\centering  
 \scalebox{0.8}{
{\footnotesize
\begin{forest}
postings/.style={
        draw=black,
        text height=3ex,
        text width=5ex,
        text depth=.25ex,
        text centered,
        rectangle
    },
for tree={circle,draw, l sep=0.2cm, s sep=0.005cm, sn edges}
[Root
  [H
      [U  
        [S 
          [S
            [A
              [I
                [N, red [ 5 , postings]]
              ]
            ]
            [E
              [I
                [N, red [ 6 , postings]]
              ]
            ]
          ]
        ]
      ] 
  ]
  [O
    [Y
      [A , red [ 2 , postings]]
    ]
    [Z
      [D
        [E
          [M
            [I
              [R, red [ 1 , postings]]
            ]
          ]
          [N, red [ 6 , postings]]
        ]
      ]
    ]
  ]
  [B
    [U
      [D
        [U
          [R, red [ 2-3-4; , postings, text width=7ex]]
        ]
        [A
        	[K, red [ 5 , postings]]
      	]
      ]
    ]
    [E
      [R
        [K, red [ 5 , postings]
          [I
            [N, red [ 1 , postings]]
          ]
        ]
      ]
    ]
  ]
  [C
  	[A
      [N, red [ 6 , postings]
        [B
          [E
            [R
              [K, red [ 1 , postings]]
            ]
          ]
        ]
      ]
    ]
  ]
  [E
    [M
      [R
        [A
          [H, red [ 4 , postings]]
        ]
        [E,red [ 2 , postings]]
      ]
    ]
  ]
]
\end{forest}
}
}
 \caption[Sample Trie]{Sample trie for sample names in Table \ref{table:names}}                
 \label{figure:sample_trie}
\end{figure*}

Figure \ref{figure:sample_trie} (positioned at the very end of the paper) is a sample trie index which was generated based on the sample records given in Table \ref{table:names}.
Each node refers to a substring of a token in reference dataset.  The \textit{red node} indicates end of token. The numbers in boxed nodes leaning from red nodes are posting lists of the tokens.  The list of record tokens was obtained by tokenizing the asciified lower case form of the record name.

\begin{table}
\begin{center}
\begin{tabular}{rl}
  User Id & Names  \\
  \hline
  1 & CANBERK BERKIN OZDEMIR   \\
  2 & AHMET EMRE BUDUR   \\
  3 & OYA CIMEN BUDUR   \\
  4 & EMRAH BUDUR   \\
  5 & HUSSAIN BERK BUDAK  \\
  6 & HUSSEIN OZDEN CAN \\
  \hline
  \end{tabular}
\end{center}
  \caption{Sample names}
  \label{table:names}
\end{table}

\subsection{Query Processing} 
In this step, we preprocessed the input raw query with two main steps.  As the first step, we have tokenized the asciified lower case form of the query and obtained a list of query tokens. Then, we expanded the list of query tokens with sequentially combined windows of existing terms.  For example, given a query \textit{"nether lands company"} in which a white space characters was introduced due to a misspelling or a possible line feed, we expanded the query as \textit{"nether lands company netherlands landscompany netherlandscompany"}.  This action prevents false negative matches of the queries having extra whitespaces injected into query terms.  On the other hand, the generated nonsense terms are eliminated by having no match in searching phase. We decided to proceed with a window of up to 4 terms as a result of our empirical analysis.

\subsection{Searching}

In this step, we traverse the trie index for each query token $t_q$ while collecting the candidate match set $M$.
One of the crucial part of our application is to apply weighted edit distance while traversing the trie index which is an expanded form of the function \textit{DFT-LOOK-UP}$_{ed}$ suggested by Arslan and Egecioglu  \cite{ARSLAN2004DictionaryDistance}.  Below are some highlights of the improvements that are introduced in the expanded form of the function. 

The original algorithm suggested by Arslan and Egecioglu was a function that returns the minimum edit distance when the query token is compared against any of the record token \cite{ARSLAN2004DictionaryDistance}.  We have improved the algorithm to collect the posting lists of the candidate record tokens having up to a given amount of edit distance from the query token.

In addition, we incorporated the confusion matrices, namely $IUC, DUC, SUC$, into edit distance calculation steps.  In this way, we were able to distinguish the unlikely edit errors from likely edit errors. For example,  we wanted to eliminate the false positive match of the query term "Saydam" with the record term "Saddam" while collecting the candidate match of the query term "Hossein" with the record term "Hussein".  Hence, by means of weigthed edit distance we were able to minimize the false positives.

The resulting algorithm is named as \textit{GT\_FreeText} and presented in Algorithm \ref{alg:search_algorithm}.  Let's first review some additional notations that is used in Algorithm \ref{alg:search_algorithm}, below.

\subsubsection{Notations for Algorithm \ref{alg:search_algorithm}} 

Let $t_{qj}$ refers to the $j$'th letter of the token $t_{q}$. 

Let $v$ denote a vertex in a tree and $v_0$ denote the root of the tree. Let the $T_c(v_p)$ be a function that enumerates the children of the parent vertex $v_p$ where $v_c \in T_c(v_p)$.

Let $T_p(v)$ be a function that returns the parent of the vertex $v$.  Let $Letter(v)$ be a function that returns the letter that corresponds to the vertex $v$.

Let $P(v)$ be a function that enumerates the posting list that corresponds to the vertex $v$.

Let the polymorphic functions  $Token(v_{\textit{eow}})$ and $Token(m_d)$ returns the record term $t_d$ that corresponds to the end of word vertex $v_\textit{eow}$ and to the match $m_d$.  

Let the polymorphic functions $Tokens(d)$ and $Tokens(M_d)$ enumerates all of the record terms in the document $d$ and the match set $M_d$ while $\|Tokens(d)\|$ and $\|Tokens(M_d)\|$ denotes the number of the record terms in document $d$ and the match set $M_d$ respectively. 

Let $m_{(d,i)}$ refer to the $i$'th match in the match set $M_d$. Let $t_{(d,i)}$ refers to the $i$'th record term in the document $d$.

Let $EOW(v)$ be a function that returns a boolean value that represent if the vertex $v$ is an end of word vertex or not.

Let IUC$_{(a,b)}$ denotes a \textit{weighted insertion unit cost} of the letter $a$ after the letter $b$ where $0 \leq$ IUC$_{(a,b)}$ $\leq 1$.  Let DUC$_{(a,b)}$ denote a \textit{weighted deletion unit cost} of the letter $a$ after the letter $b$ where $0 \leq$ DUC$_{(a,b)}$ $\leq 1$. Let SUC$_{(a,b)}$ denote a \textit{weighted substitution unit cost} of the letter $a$ for the letter $b$ where $0 \leq$ SUC$_{(a,b)}$ $\leq 1$.  IUC$_{(a,b)}$ = DUC$_{(a,b)}$ = SUC$_{(a,b)} = 1$ where either of the letters $a$ or $b$ is a non-ascii character including NULL.

\begin{algorithm}
\caption{\textit{GT\_FreeText ALGORITHM}}
\label{alg:search_algorithm}
\begin{algorithmic}[1]
\Procedure{GT\_FreeText}{$q$} 
\State $M \gets \textit{init}$ \Comment{Initialize candidate set dictionary}
\For{all $t_q \in q$}
\State $\textit{TRAVERSE\_TRIE$_1$}(t_q, M)$
\EndFor
\State \Return M
\EndProcedure
\\
\\
\Procedure{\textit{TRAVERSE\_TRIE$_1$}}{$t_q, M$} 
\State $m = \textit{len}(t_q)$ \Comment{Length of query term}
\State $\epsilon \gets \tau_m$
\State $l \gets 1$ \Comment{Init level to 1}
\For{all $v_c$ where $v_c\in T_C(v_0)$}
\State $\textit{TRAVERSE\_TRIE$_2$}(v_c, M, \epsilon, l, t_q)$
\EndFor
\EndProcedure
\\
\\
\Procedure{\textit{TRAVERSE\_TRIE$_2$}}{$v_p, M, \epsilon, l, t_q$} 

\State $v_{pp} \gets T_p(v_p)$
\State $ch_{da} \gets Letter(v_{pp})$ \Comment{Previous letter of $t_d$}
\State $ch_{db} \gets Letter(v_{p})$ \Comment{Current letter of $t_d$}
\State $ch_{qa} \gets NULL$ \Comment{Previous letter of $t_q$}

\State $m = \textit{len}(t_q)$ \Comment{Length of query term}
\For{$j = 0$ to $m$}
    \State $ch_{qb} \gets t_{qj}$ \Comment{Current letter of $t_q$}
    \State IC $\gets D_{v,i,j-1} +   IUC_{(ch_{da}, ch_{db})} $ 
    \State DC $\gets D_{v,i-1,j} +   DUC_{(ch_{qa}, ch_{qb})} $ 
    \State SC $\gets D_{v,i-1,j-1} +  SUC_{(ch_{qb}, ch_{db})}$ 
	
    \State $D_{u,i,j} \gets \min \{$ IC, DC, SC $\}$
    \State $ch_{qa} \gets ch_{qb}$
\EndFor

\State $e \gets D_{u,l,m}$ 

\If {$EOW(v_p) == \textit{true}$ AND  $e \leq \epsilon$}
    \For{$d \in P(v_p)$} \Comment{Collect posting list}
        \State $t_d \gets Token(v_p)$
        \State $m_d \gets (t_q, t_d, e)$
        \State Append $m_d$ to $M_d$
    \EndFor
\EndIf

\State min\_distance $\gets \min \{D_{u,l,j} | 0 \leq j \leq m \}$ 

\If{min\_distance $\leq \epsilon$}
	\State $l \gets l + 1$ 
    \State $\epsilon \gets \tau_{\max(m, l)}$
    \For{all $v_c$ where $v_c\in T_c(v_p)$} 
        \State $\textit{TRAVERSE\_TRIE$_2$}(v_c, M, \epsilon, l)$
    \EndFor
\EndIf

\EndProcedure

\end{algorithmic}
\end{algorithm}

\subsection{Filtering} 
The aim of the filtering step is to improve the quality and performance of the ranking step. We believed that the best way to rank is not to rank.  In other words, we will have a  better ranking step if we filter out the irrelevant matches that does not deserve to be ranked. In this section, we will introduce some tips of the filtering step.

For the first tip, let's take the example given in Table \ref{table:filter_noisy_words}.  In this example a query token CORPORATION caused many records that contain this token.  Considering that there are possibly even more records that contain this token which was shown in Table \ref{table:doc_frequencies}, presenting these matches as candidate match will fill up the top $k$ slots with these nonsense matches.  As a result, it will prevent true positive matches to take a significant slot in top $k$ result set. Since our aim is to improve the quality of the matches in top $k$ slots we filtered out those results that consist of a single match $m_d = (t_q, t_d, e)$ where $DocFreq(t_d) > k$

As an example for the second tip, we have analyzed  another frequent token "GLOBAL" which co-occurs with the term "CORPORATION" as shown in Table \ref{table:filter_frequent_item_sets}. Since the number of documents that contain both of the terms GLOBAL and CORPORATION is still more than 100, which was also shown in Table \ref{table:frequent_item_sets}, presenting these matches will fill up top $k > 100$ slots unless we eliminate them.  In order to eliminate them, we keep track of the number of records that corresponds to the candidate sets $M_d$ and filter out those match sets $M_d$ where $Support(M_d) \geq k$.

As the final tip, we want to mention about the unique records whose individual terms are all frequent terms as shown in Table \ref{table:keep_unique_term_sets}. If we have a match set $M_d$ where $Support(M_d) \leq k$ we let them take a slot in top $k$ candidate match set.

\begin{table}
\begin{center}
\begin{tabular}{rl}
  TERMS & DOC FREQUENCY\\
  \hline
  CORPORATION & 7000+   \\
  BANK & 6000+   \\
  INTERNATIONAL & 5000+   \\
  SECURITIES & 3000+   \\
  GLOBAL & 2000+  \\
  \hline
  \end{tabular}
\end{center}
  \caption{Document frequencies of sample noisy words}
  \label{table:doc_frequencies}
\end{table}

\begin{table}
\begin{center}
\begin{tabular}{rl}
  TERMS & SUPPORT\\
  \hline
  CORPORATION + INTERNATIONAL & 2000+   \\
  GLOBAL + CORPORATION & 100+   \\
  CORPORATION + SECURITIES & 40+   \\
  \hline
  \end{tabular}
\end{center}

  \caption{Support of frequent item sets}
   \label{table:frequent_item_sets}
\end{table}

\begin{table}
\begin{center}
\begin{tabular}{rl}
  Query & INNOCENTA \textbf{CORPORATION}  \\
  \hline
  Resultset &  \\
  \hline
  1 & BADDY \textbf{CORPORATION}   \\
  2 & WANTED INTL \textbf{CORPORATION}   \\
  3 & SANCTIONED \textbf{CORPORATION} LTD   \\
  4 & BOMBER \textbf{CORPORATION}   \\
  5 & NARCOTIC \textbf{CORPORATION}  \\
  \hline
  \end{tabular}
\end{center}
  \caption{Filtering out noisy word matches}
  \label{table:filter_noisy_words}
\end{table}

\begin{table}
\begin{center}
\begin{tabular}{rl}
  Query & INNOCENTA \textbf{GLOBAL CORPORATION}  \\
  \hline
  Resultset &  \\
  \hline
  1 & BADDY \textbf{GLOBAL CORPORATION}   \\
  2 & WANTED INTL \textbf{GLOBAL CORPORATION}   \\
  3 & SANCTIONED \textbf{GLOBAL CORPORATION} LTD   \\
  4 & BOMBER \textbf{GLOBAL CORPORATION}   \\
  5 & NARCOTIC \textbf{GLOBAL CORPORATION}  \\
  \hline
  \end{tabular}
\end{center}
  \caption{Filtering out frequent itemsets}
    \label{table:filter_frequent_item_sets}
\end{table}

\begin{table}
\begin{center}
\begin{tabular}{rl}
  RECORDS & SUPPORT \\
  \hline
  GLOBAL CORPORATION SECURITIES & 1   \\
  BANK INTERNATIONAL & 1   \\
  INTERNATIONAL CORPORATION BANK & 1   \\
  \hline
  \end{tabular}
\end{center}
  \caption{Unique term sets}
    \label{table:keep_unique_term_sets}
\end{table}

\subsection{Ranking} 

In this step, we need to calculate a score for each records in the match sets $M$, which is given out of the filtering step.  Contrary to the ordinary scoring schemes commonly adopted by the mainstream search engines, the resulting scores of this step must reflect percentage similarity of the record name compared to the matched query terms. In other word, a score of 100 will refer to an exact match while the score of 0 will be given to no match at all.  After calculating percentage scores, we sort the matches  descendingly by their percentage scores and select top $k$ candidate results.  Below is step by step formulation of this process.

Given $m_d = (t_q, t_d, e)$, let the function 

\begin{equation}
MI(m_d)=Sim(t_q, t_d) \times I(t_d)
\end{equation}

gives the mutual information of the match $m_d$ while  

\begin{equation}
MI^w(m_d)=Sim^w(t_q, t_d) \times I(t_d)
\end{equation}

returns the weigthed mutual information of the match $m_d$. 

Then, the total mutual information $TMI_d$ of the match set $M_d$ for the document $d$ is defined as follows:

\begin{equation}
TMI_d (M_d) = \sum_{i=1} ^ {\|M_d\|} MI(m_{(d,i)})
\end{equation}

On the other hand, the total weighted mutual information $TMI^w_d$ of the match set $M_d$ for the document $d$ is defined as follows:

\begin{equation}
TMI_d^w (M_d) = \sum_{i=1} ^ {\|M_d\|} MI^w(m_{(d,i)})
\end{equation}

The total information in the document $d$ is defined as follows:

\begin{equation}
TID_d = \sum_{i=1} ^ {\|Tokens(d)\|} I(t_{(d,i)})
\end{equation}

So, the percentage score of the document $d$ that corresponds to a particular match set $M_d$ is defined as follows:

\begin{equation}
Score (d, M_d) = 100 \times \frac{TMI(M_d)}{TID(d)}
\label{equation:weighted_score}
\end{equation}

And the weighted percentage score of the document $d$ that corresponds to a particular match set $M_d$ is defined as follows:

\begin{equation}
Score^w (d, M_d)= 100 \times \frac{TMI^w(M_d)}{TID(d)}
\label{equation:weighted_score}
\end{equation}

Finally, in the ranking step each of the match set $M_d \in M$ is scored by means of the functions given in Eq.\ref{equation:weighted_score} and ranked descendingly by this score.  As a result, the top $k$ candidate result set is obtained.

\subsection{Scaling} 

In this part of our application, we aimed to revise the architectural design so that our application can be safely scaled out while increasing the number of records it can search from.

We splitted our target reference records into $n$ segments of $r$ records and created multiple trie data structure for each segment. As a result, we obtained a forest of trie data structures each containing $r$ records.  

\begin{figure}[hb]
  \centering
  \includegraphics[width=3.5in]{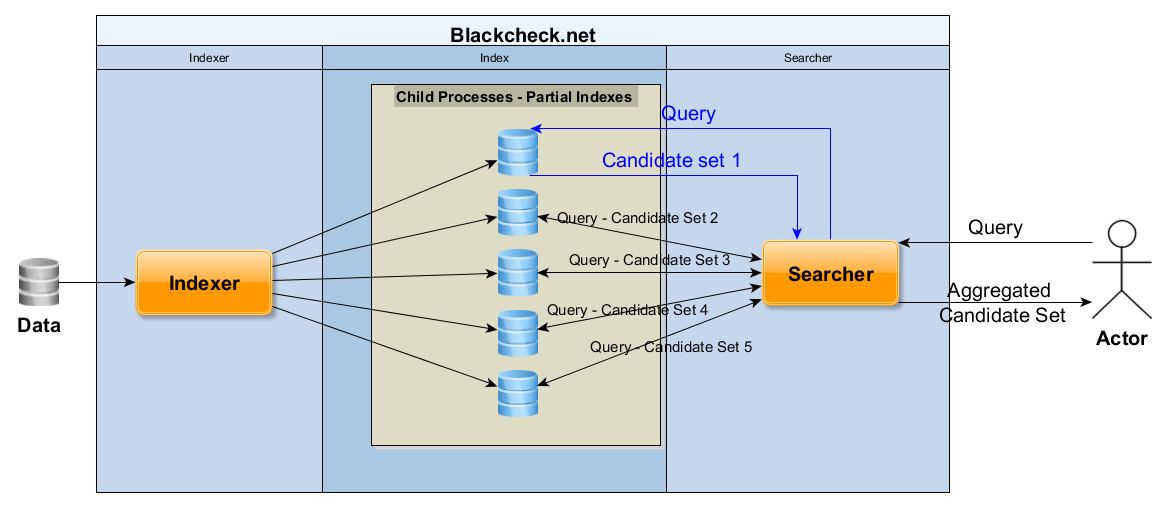}
  \caption[Scaling out architecture of the application]
   {Scaling out architecture of the application}
  \label{figure:scale_out}
\end{figure}

In the query time, we searched the query $q$ from each trie data structure and obtained top $k$ candidate results from each trie.  At the end, we merged $n$ candidate result sets and sorted all results by the calculated scores and returned the top $k$ records from the resulting merged and aggregated result sets. Figure \ref{figure:scale_out} shows the resulting architectural diagram. 

Each of trie index can be served by a seperate process which can be running either all in the same machine or distributed machines.

In this way, we are able to scale out the application while increasing the number of the records in the target dictionary and preserving the capability of addressing all of the constraints of the problem.

\section{EXPERIMENTS}
We carried out a series of experiments on a labeled dataset to figure out the performance of the algorithm in terms of response time, indexing time and response quality.  Below are the details of each type of analysis along with the details of the dataset.

\subsection{Dataset}
We have collected two main type of data sets such as queries datasets and reference datasets.

\subsubsection{Queries Datasets}
We collected three different dataset from a leading bank in Turkey such as structural individual queries $(Q_{in})$, structural corporate queries $(Q_{co})$ and unstructured free text queries $(Q_{mix})$.  The details of the datasets are described below. 

\subsubsection*{Structured Individual-Typed Queries $(Q_{in})$}
The structured individual dataset consists of full names of individuals.  The total number of queries in this dataset is 2110502.  These queries are searched from a list of individual names of size 2038234.  The dataset is not labeled and it was used just to test the response time performance of the applications in individual-typed queries.

\subsubsection*{Structured Corporate-Typed Queries $(Q_{co})$}
The structured corporate dataset consists of the legal names of corporates.  The total number of queries in this dataset is 488803. These queries are searched from a list of reference corporate names of size 207468. The query dataset is not labeled and it was used just to test the response time characteristics of the applications in corporate-typed queries.

\subsubsection*{Mixed Unstructured Free-Text Queries $(Q_{mix})$}
The mixed unstructured free-text queries dataset consists of a fraction of randomly selected  international money transfer queries that are received in one-month time frame.  The total number of the queries in the dataset are 406928.  But we discovered that many queries are redundant thus we aggragated the dataset. As a result, the number of distinct queries in the dataset turned out to be 85572. All of the distinct queries are either labeled as true positive match with certain record name or true negative match. The number of queries that are flagged as true positive flag with at least one record text is 8409.  On the other hand a total of 12272 record names have been flagged as a true positive match with a certain query.

Table \ref{table:sample_dataset} shows a sample snapshot of the labeled dataset.  Note that the query \textit{"435021 BANK KBC"} is flagged as true positive match with two different record texts. Note also that the query text \textit{"INVOICE RECEIPT"} is flagged as a true negative match since it has no corresponding matching record text.

\subsubsection{Reference Datasets}
We used three different reference datasets to search from.
\subsubsection*{Individual Entities $(R_{in})$}
This dataset consists of 2038234 entities of individual names.
\subsubsection*{Corporate Entities $(R_{co})$}
This dataset consists of 207468 entities of corporate legal names.
\subsubsection*{Small Mixed Entities $(R_{mix})$}
This dataset consists of 43019 entities of both individual and corporate legal names.

\subsection{Applications}
We have benchmarked 2 different applications under the experimentation phase.  Below are the brief details of these applications.

\subsubsection{GT\_FreeText} 
The application framework that is presented in this paper is named as \textit{GT\_FreeText} throughout the experimental analysis.

\subsubsection{LSH} 
We have benchmarked our application against a locality sensitive hashing framework that is provided by Informatica, namely Name3.    The configuration of the application was done by a local representative of the application vendor.  The line of business application that we benchmarked in this experimentation phase stores the hash indexes in a relational database rather than in-memory. This application was named as \textit{LSH} throughout the experimental analysis.

\begin{table}
\begin{center}
\resizebox{\columnwidth}{!}{%
\begin{tabular}{lll}
$Q_{Type}$  & Query Text &  Record Text  \\
  \hline
$in$ &     \textbf{MARIA CELTIQ} &  \textbf{MARIAN} OYA \textbf{CELTIK} \\ 
$in$ &     \textbf{AHMET} EMRE \textbf{MIYESE} &  \textbf{AHMET MIYESE} \\ 
$co$   &       \textbf{CITI BANK} &  \textbf{CITY BANK}  \\
$co$    &  \textbf{HERMANN} NIMCOM GMBH &  \textbf{HERMANN}  \\
$co$ &       DALGA\textbf{DURAN MAKIN}A A.S.
       &  \textbf{DURAN MAKIN}  \\
       
$mix$ &      \textbf{MUHAMMED SALIH} AHMET EMRE &  \textbf{Muhammad SALAH}  \\
$mix$    &      \textbf{ATC} ENTERPRISES \textbf{LTD} KAYSERI &  \textbf{ATC LTD}  \\

$mix$   &         435021 BANK \textbf{KBC} &  \textbf{KWANGSON BANKING CO.}  \\
&  &  \textbf{KBC} FINANCIAL INC  \\
$mix$    & \textbf{ODESSA} &  \textbf{ODESSA} AIR  \\
$mix$ & INVOICE RECEIPT &   \\ 
  \hline
  \end{tabular}
  }
\end{center}
  \caption{Sample dataset}
    \label{table:sample_dataset}
\end{table}

\subsection{Evaluation}

We have evaluated two main characteristics of the application, namely temporal characteristics and the quality of result set.  The former is the temporal analysis in which we measured the indexing time and response time of the application.  The latter measures the quality of the result sets of each applications compared to the human evaluations.

\subsection{Temporal Anaysis}
Indexing times and response times of the applications were analyzed under this section. 

\subsubsection{Indexing Time Anaysis}
\begin{figure}[h!]
  \centering
  \includegraphics[width=3.5in]{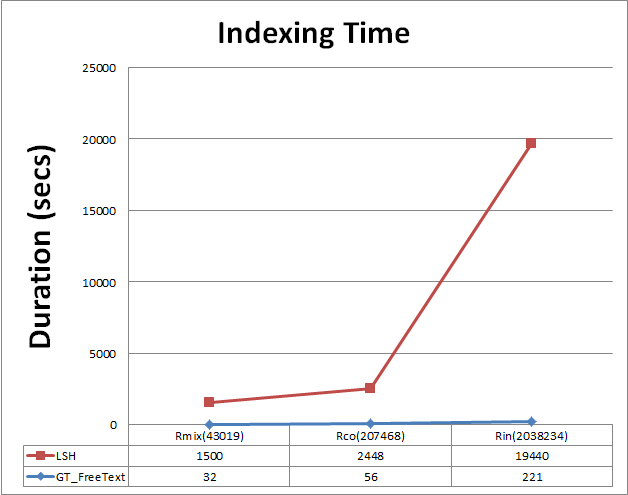}
  \caption[Indexing Time]
   {Indexing Time}
  \label{fig:indexing_time}
\end{figure}

Each of three reference datasets were indexed by using GT\_FreeText and LSH.  The resulting indexing time for each application was shown in Fig \ref{fig:indexing_time}. It can be observed that the indexing time complexity of GT\_FreeText is sublinear while LSH presents exponential time complexity. Considering that LSH stores its hash indexes in a relational database, we may expect better indexing time for LSH if it would store the indexes in-memory.
Nevertheless, we can safely conclude that GT\_FreeText can index millions of records in a few minutes.

\subsubsection{Response Time Anaysis}

\begin{figure}[!tbp]
\begin{subfigure}[b]{0.2\textwidth}
    \includegraphics[width=\textwidth]{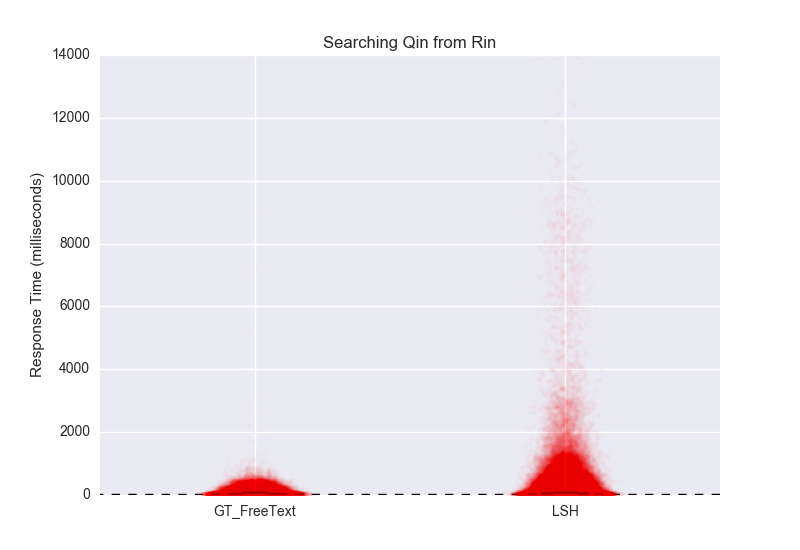}
    \caption{Scattered datapoints}
    \label{figure:response_time_in_1}
  \end{subfigure}
  \hfill
  \begin{subfigure}[b]{0.2\textwidth}
   \includegraphics[width=\textwidth]{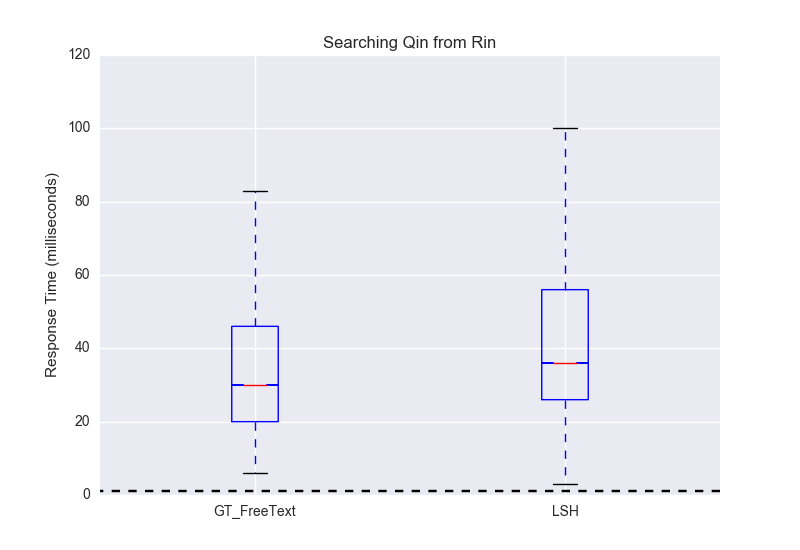}
    \caption{Closer look}
    \label{figure:response_time_in_2}
  \end{subfigure}
  \caption{Response time of searching $Q_{in}$ from $R_{in}$}
  \label{figure:response_time_in}
\end{figure}

\begin{figure}[!tbp]
\begin{subfigure}[b]{0.2\textwidth}
    \includegraphics[width=\textwidth]{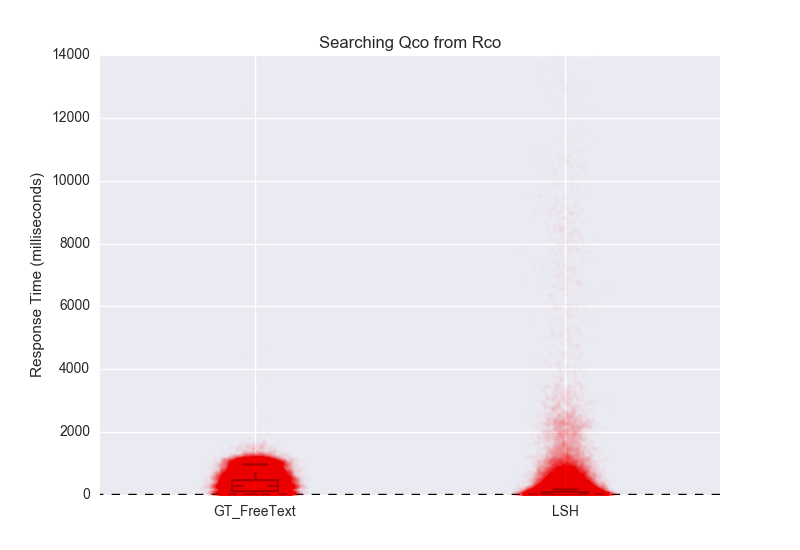}
    \caption{Scattered datapoints}
    \label{fig:response_time_co_1}
  \end{subfigure}
  \hfill
  \begin{subfigure}[b]{0.2\textwidth}
   \includegraphics[width=\textwidth]{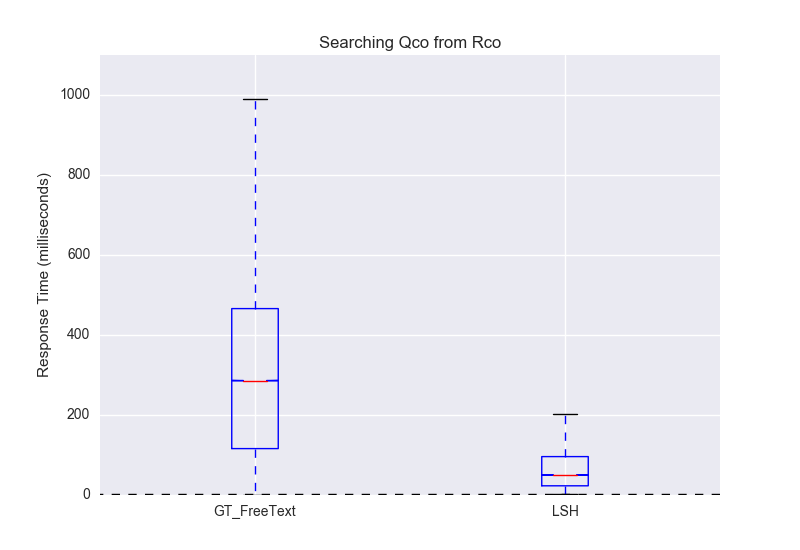}
    \caption{Closer look}
    \label{fig:response_time_co_2}
  \end{subfigure}
  \caption{Response time of searching $Q_{co}$ from $R_{co}$}
    \label{figure:response_time_co}
\end{figure}

\begin{figure}[!tbp]
\begin{subfigure}[b]{0.2\textwidth}
    \includegraphics[width=\textwidth]{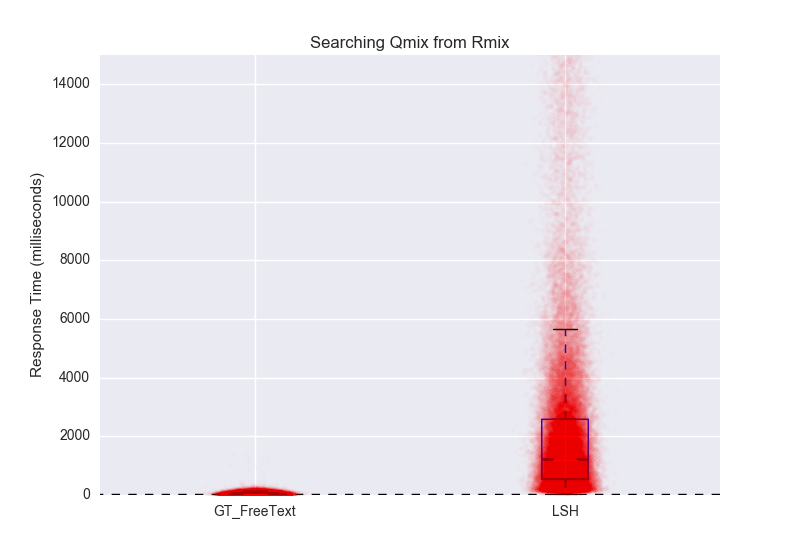}
    \caption{Scattered datapoints}
    \label{fig:response_time_unstructured_1}
  \end{subfigure}
  \hfill
  \begin{subfigure}[b]{0.2\textwidth}
   \includegraphics[width=\textwidth]{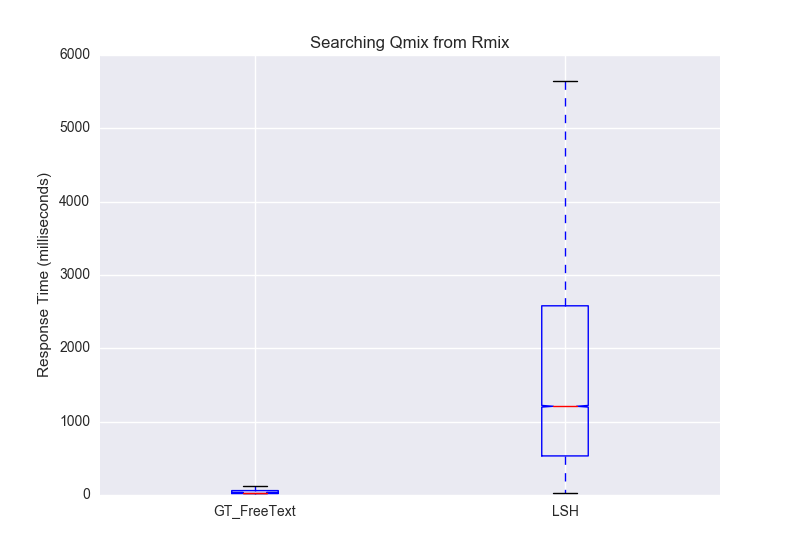}
    \caption{Closer look}
    \label{fig:response_time_unstructured_2}
  \end{subfigure}
  \caption{Response time of searching $Q_{mix}$ from $R_{mix}$}
    \label{figure:response_time_unstructured}
\end{figure}

Response times of the applications GT\_FreeText and LSH are analyzed across three different query datasets, namely structured individual-typed queries, structured corporate-typed queries and unstructured mixed-typed queries.  The resulting response times were presented in Figure \ref{figure:response_time_in},\ref{figure:response_time_co} and \ref{figure:response_time_unstructured} respectively.
The scattered data points for each plot shows that the response times to each type of query is way more lower in GT\_FreeText compared to LSH. 

Closer look at the box plot of individual-typed structural queries shows that response time of the majority of the queries remains comparable for GT\_FreeText and LSH.  However, the scattered plot reveals that many queries takes more than 2 seconds in LSH  which is unacceptable for real time financial transaction processing context.   The scattered plot gives a clear picture to conclude that the response time of individual-typed queries in GT\_FreeText application remains under 1 second.

The boxplot of corporate-typed queries shows that the latency of GT\_FreeText for the majority of the corporate-typed queries is higher than that of LSH.  However, the scattered plot reveals that GT\_FreeText returns its response within 2 seconds where LSH fails to respond in 2 seconds for many queries. 

LSH turned in its worst response time performance in mixed-typed unstructured free-text queries which was projected on both the box plot and scattered data points in Figure \ref{fig:response_time_unstructured_1} and \ref{fig:response_time_unstructured_2} respectively. It can be easily observed that GT\_FreeText preserved its compelling response time performance also for mixed-type unstructured queries.

\subsection{Quality Anaysis}
\begin{figure}[h!]
  \centering
  \includegraphics[width=3.5in]{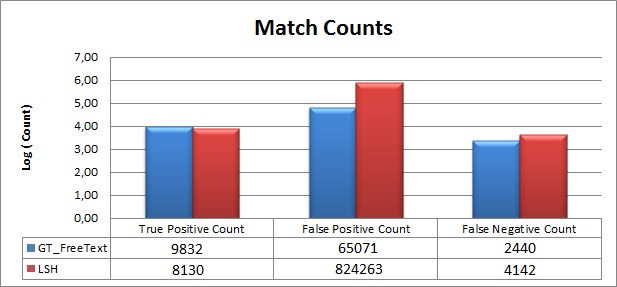}
  \caption[Match Counts]
   {Match Counts}
  \label{figure:match_counts}
\end{figure}

\begin{figure}[h!]
  \centering
  \includegraphics[width=3.5in]{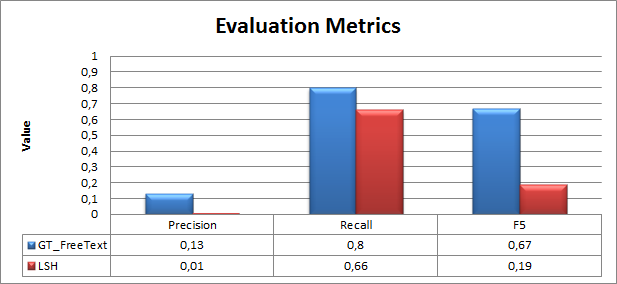}
  \caption[Evaluation Metrics]
   {Evaluation metrics}
  \label{figure:evaluation}
\end{figure}

Controlled experimentation has been conducted on the unstructured free-text queries dataset in which true positive matches were labeled.  The dataset were queried across the applications GT\_FreeText and LSH.  Fig \ref{figure:match_counts} and \ref{figure:evaluation} shows the evaluation metrics along with the match counts that were calculated for the result set of each application.  Below is the comparative analysis of each type of metrics.

\subsubsection{Precision}
The ratio of the relevant documents retrieved to the total number of retrieved documents.  
\begin{equation}
\textrm{precision} = 
\frac{
	\textrm{\# of relevant documents retrieved}
}
{
	\textrm{\# of retrieved documents}
}
\end{equation}

The true positive match count of GT\_FreeText is greater than the one in LSH.  On the other hand, LSH  returned way more false positive results.  Hence the precision of GT\_FreeText turned out to be greater than LSH as seen in Figure \ref{figure:evaluation}.

\subsubsection{Recall}
The ratio of the relevant documents retrieved to the total number of relevant documents.
\begin{equation}
\textrm{recall} = 
\frac{
	\textrm{\# of relevant documents retrieved}
}
{
	\textrm{\# of all relevant documents}
}
\end{equation}
Higher true positive match count and lower false negative error count of GT\_FreeText made its recall value greater than LSH as shown in Figure \ref{figure:evaluation}. 

\subsubsection{F-Measure}
The weighted harmonic mean of precision and recall.  It can be formulated as 
\begin{equation}
\textrm{F}_\beta = (1+\beta^2) \frac{\textrm{precision} \times \textrm{recall}}
{\beta^2 \times \textrm{precision} + \textrm{recall}}
\end{equation}
where $\beta$ adjusts the importance of precision and recall relative to each other.  We used $F_5$ for evaluation since recall is way more important than precision and since extraction recall is essential to address \textit{avoiding false negative} constraint.

Since the cost of false negative matches is prohibitively higher than the cost of false positive match in the context of terrorism financing, we penalized false negative matches 5 times more than false positive matches by using $F_\beta$ measure where $\beta = 5$.  We decided the value of $\beta = 5$ as a
result of our empirical analysis in which we observed that the function $F_\beta$ doesn't reflect any significant change when $\beta > 5$.

As a result, since GT\_FreeText outperformed in terms of both higher true positive matches and lower false negative error, therefore the value of $F_5$ in GT\_FreeText turned out to be greater than the one of LSH as given in Figure \ref{figure:evaluation}.

\section{CONCLUSIONS}
The need for screening financial transaction against terrorism entities is increasing at a rapid race as it is being enforced by governmental institutions including US-OFAC.  Financial institutions have to administer a real-time fast and scalable solution for this problem. Inspite of its critical importance, this problem has gone unseen in the field of Information Retrieval. We have proposed a solution for this problem addressing various constraints defined by the relevant business experts of a leading bank in Turkey. The building blocks of such solution is analysed such as \textit{query processing, searching, filtering and ranking} by presenting a number of useful tips and tricks.  The performance of the final application is evaluated in terms of indexing time, response time and the quality of the results by comparing with a line of business application which was based on locality sensitive hashing.  The results clearly shows that the proposed solution addresses all of the predefined constraints while outperforming the benchmarked application not only for structured queries but unstructured free-text queries also.

\addtolength{\textheight}{-12cm}   




\section*{ACKNOWLEDGMENT}

First and foremost, the author needs to say that formal thanks are inadequate to express his gratitude to his beloved wife Oya Cimen Budur who is always a powerful source of energy for him with her consistent support to keep him on the right track.

The author offers his sincerest gratitude to Canberk Berkin Ozdemir for his continous support and substantial contribution on this project.

The author is most indebted to
Gokcer Belgusen and Tolga Yavuz who spared no effort to achieve outstanding results on this study.

The author owes a duty of good faith and fidelity towards his employer Garanti Technology who let him investigate every aspects of the problem to come up with a competitive solution while giving him the freedom to accomodate his personal priorities.

The author is glad to thank his dear co-workers particularly Aybuke Kurues Seyitogullari, Cemre Yalvac, Zeynep Hiz, Elcin Ugurlu Kasap,  Engin Sag and last but not least Mustafa Duman since the author was previledged to have generous support and insightful comments from them in all part of this study.


\bibliography{Mendeley} 
\bibliographystyle{unsrt}
\vfill

\end{document}